\documentclass[10pt,twocolumn,letterpaper]{article}

\usepackage{cvpr}              

\usepackage{color}
\usepackage{colortbl} 

\definecolor{mygreen}{rgb}{0.3, 0.9, 0.4}
\definecolor{BurntOrange}{rgb}{1.0, 0.35, 0.0}

\newcommand{\blue}[1]{\textcolor{black}{#1}}

\newenvironment{blueenv}{\color{black}}{}
\usepackage{amsmath}
\usepackage{amssymb}
\usepackage{graphicx}
\usepackage{multirow}

\usepackage{algorithm}
\usepackage{algorithmicx}
\usepackage{booktabs}
\usepackage{algpseudocode}  
\usepackage{subcaption}
\usepackage{xspace}

\usepackage{tikz}
\usetikzlibrary{arrows.meta, decorations.pathreplacing, calc}

\definecolor{cvprblue}{rgb}{0.21,0.49,0.74}
\usepackage[pagebackref,breaklinks,colorlinks,allcolors=cvprblue]{hyperref}

\def\paperID{12} 
\def\confName{CVPR}
\def\confYear{2026}

\author{
Michele De Vita$^{1}$ \quad 
Julian Wiederer$^{2}$ \quad 
Vasileios Belagiannis$^{1}$ \\[6pt]
$^{1}$Friedrich-Alexander-Universität Erlangen-Nürnberg, Germany \\
$^{2}$Mercedes-Benz AG, Germany\\[2pt]
{\tt\small \{first.last\}@fau.de} \quad {\tt\small julian.wiederer@mercedes-benz.com}
}

\usepackage{siunitx}

\title{Forecasting the Past: Gradient-Based Distribution Shift Detection in Trajectory Prediction}

\begin{document}

\maketitle
\thispagestyle{empty}
\pagestyle{empty}


\begin{abstract}
Trajectory prediction models often fail in real-world automated driving due to distributional shifts between training and test conditions.
Such distributional shifts, whether behavioural or environmental, pose a critical risk by causing the model to make incorrect forecasts in unfamiliar situations.
We propose a self-supervised method that trains a decoder in a post-hoc fashion on the self-supervised task of forecasting the second half of observed trajectories from the first half. 
\blue{The L2 norm of the gradient of this forecasting loss with respect to the decoder's final layer defines a score to identify distribution shifts.
Our approach, first, does not affect the trajectory prediction model, ensuring no interference with original prediction performance and second, demonstrates substantial improvements on distribution shift detection for trajectory prediction on the Shifts and Argoverse datasets.
Moreover, we show that this method can also be used to early detect collisions of a deep Q-Network motion planner in the Highway simulator. 
Source code is available at \footnote{\url{https://github.com/Michedev/forecasting-the-past}}.}
\end{abstract}


\section{Introduction}
\label{sec:introduction}

\label{par:problem_definition}

\begin{figure}[t]
  \centering
  \begin{subfigure}{0.48\textwidth}
      \includegraphics[width=\linewidth, clip, trim={0.5cm 7.6cm 0.5cm 3.3cm}]{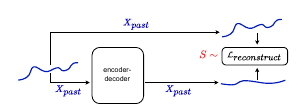}
      \caption{Previous reconstruction-based methods.}
      \label{fig:teaser_reconstruct}
  \end{subfigure}
  \begin{subfigure}{0.48\textwidth}
      \includegraphics[width=\linewidth, clip]{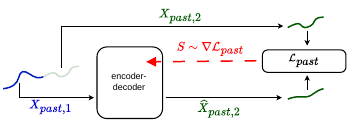}
      \caption{Our gradient-based method.}
      \label{fig:teaser_ours}
  \end{subfigure}
  \caption{Instead of (a) reconstructing the entire past trajectory $X_{past}$ and using the reconstruction loss $\mathcal{L}_{reconstruct}$ as a distribution shift score, we (b) forecast a portion of the observed trajectory $X_{past}$ and use the gradient of the forecasting loss, $\nabla \mathcal{L}_{forecast}$, as the distribution shift score $S$. Specifically, we divide $X_{past}$ into $X_{past,1}$ and $X_{past,2}$ and forecast $\hat{X}_{past,2}$ given $X_{past,1}$. The gradient $\nabla \mathcal{L}_{forecast}$ effectively identifies unknown scenarios in trajectory prediction.}
  \label{fig:teaser}
\end{figure}

Forecasting the movements of surrounding entities is critical for safe automated driving. 
However, state-of-the-art models often fail in real-world settings due to natural distribution shifts between training and test conditions.
These shifts can occur in various forms: spatial, \eg, vehicles in unexpected locations, behavioural, \eg, erratic movements and traffic violations, and environmental, \eg, new weather conditions and unseen road layouts.
Degradation in prediction performance creates severe safety risks as models may make overconfident wrong predictions in unfamiliar scenarios, which can lead to catastrophic failures during motion planning.
Nevertheless, existing trajectory prediction research has largely focused on improving prediction accuracy, often neglecting crucial aspects of model robustness and reliability.


\label{par:previous_approaches}
In trajectory prediction, detecting distribution shifts requires methods that can identify deviations in test data from training distributions, without access to ground truth labels during training. 
Prior work addresses the problem through three main paradigms: reconstruction-based methods~\cite{chakraborty2023Structural,gao2024multitransmotion,hornauer2023heatmap} that identify anomalous samples through high reconstruction error, one-class methods~\cite{MAAD2021} that learn boundaries around normal data, and uncertainty quantification approaches~\cite{wiederer2023joint,marvi2025evidential} that estimate model confidence using probabilistic estimation. 
However, these approaches have major limitations. 
Reconstruction-based methods struggle to distinguish input complexity from true anomalies. 
One-class methods require careful hyper-parameter tuning for high-dimensional trajectory data~\cite{pmlr-v164-wiederer22a}. 
Uncertainty quantification approaches depend on model calibration and often conflate different types of uncertainty~\cite{Rhinehart2020Deep,filos2020can, holle2025uncertainty}.

\label{par:key_contribution}
To address these issues, we present a post-hoc method for distribution shift detection that operates without degrading the performance of pre-trained trajectory prediction models and hyper-parameter tuning. 
\blue{Our approach leverages the observation that the trajectory prediction loss produces distinct gradients for in-distribution and distribution shifted scenarios.
However, as we don't have access of future trajectories at test time to compute this loss, we propose a self-supervised forecasting task, where a decoder predicts the second half of the historical trajectory given the first half.}
The method computes the L2 norm of the gradient of this self-supervised forecasting loss $\mathcal{L}_{\text{past}}$ with respect to the decoder final layer pre-activations as a score for distribution shift detection.
Crucially, the pre-trained encoder remains frozen, ensuring no interference with the original model predictions. 
Our experiments demonstrate substantial improvements: On the Shifts dataset~\cite{malinin2021shifts}, we achieve 71.0\% AUROC compared to 56.8\% for existing methods, while on Argoverse~\cite{Argoverse} we consistently exceed 70\% AUROC for detecting artificially removed trajectory patterns. 
\blue{Then we deploy our model in the interactive Highway simulator to monitor how our method behaves in an online reinforcement learning. The gradient-based score outperform other self-supervised approaches in detecting collisions one second it happens.}
The method successfully identifies diverse distribution shifts including environmental changes (new cities, weather conditions), behavioural anomalies (turning patterns, collisions), and velocity-based outliers.

\label{par:contributions_list}
We summarize our contributions as follows:
\begin{itemize}
    \item We present a post-hoc approach for detecting distribution shifts in pre-trained trajectory prediction models that does not affect performance.
    \item We propose a self-supervised trajectory forecasting loss function to extract the gradient norm, which is then employed as an anomaly score at test time.
    \item Our experiments demonstrate that our approach can successfully be used for out-of-distribution and anomaly detection in driving environments. On the Shifts Vehicle Motion Prediction dataset \cite{malinin2021shifts} we increase the ROC AUC score from 56\% to 71\%, showcasing the effectiveness of our method. On Argoverse~\cite{Argoverse}, the method achieves AUROC over 70\%. \blue{On the Highway simulator, the gradient-based score outperforms other self-supervised approaches on collision detection.}
\end{itemize}

\section{Related Work}
\label{sec:relworks}

\paragraph{Trajectory Prediction}
\label{par:traj_prediction_related}
Accurate trajectory prediction is essential for autonomous driving systems, as prediction errors can lead to safety-critical planning failures.
To advance performance on the benchmarks~\cite{Caesar_2020_CVPR, Chang_2019_CVPR, wilson2023argoverse2generationdatasets, feng2025UniTraj, Ettinger_2021_ICCV, interactiondataset,agents-llm}, numerous approaches have been proposed, with recent progress driven by agent- and map-aware architectures based on graph neural networks and transformers~\cite{Pazho_2024_CVPR, zhou2022hivt, wagner2024jointmotion, pmlr-v157-chen21a, liu2025dyttptrajectorypredictionnormalizationfree, Aydemir_2023_ICCV}.
For our distribution shift detection, we adopt the graph-based Hierarchical Vector Transformer (HiVT) from~\cite{zhou2022hivt} and Transformer \cite{vaswani2017attention} as the architecture for our experiments, but our proposed method can be applied to any architecture.

\paragraph{Distribution Shift Detection in Trajectory Prediction}
\label{par:dist_shift_detection_related}

Although trajectory prediction models exhibit strong performance on standard benchmarks, they often degrade under distribution shifts.
Long term established approaches broadly fall into three paradigms.
Uncertainty quantification estimates model confidence, often using ensembles~\cite{lakshminarayanan2017simple} or Monte Carlo dropout~\cite{malinin2021shifts,gal2016dropout}.
Reconstruction-based methods detect anomalies through high reconstruction error, for example with graph autoencoders~\cite{pmlr-v164-wiederer22a} or recurrent variational autoencoders~\cite{chakraborty2023Structural}.
Representation-based approaches instead model the normal data distribution in latent space, such as Gaussian mixture models~\cite{wiederer2023joint} or diffusion models~\cite{li2024difftad, yao2024trajoutofdistribution}.
While these paradigms evaluate the model output or latent space representations, they overlook the gradient magnitudes computed with respect to the network parameters. These gradients naturally distinguish in-distribution from distribution shifted scenarios without altering the pre-trained architecture.

\paragraph{Gradients for Distribution Shift Detection}
\label{par:gradients_related}
While most of the literature in distribution shifts detection focuses on loss or feature space, recent works have explored gradient-based distribution shift detection across various domains. Huang et. al.~\cite{huang2021importance} demonstrate that gradient norms can serve as effective indicators for distribution shift samples in image classification, showing that distribution shift inputs tend to produce larger gradients due to increased model uncertainty. Similarly, \cite{liang2018enhancing} propose using the magnitude of gradients computed from the softmax outputs as a confidence measure for detecting misclassified and distribution shift examples. More recently, \cite{zhang2025gradient} proposes a gradient based method to address the problem of overconfident prediction of neural network.
To the best of our knowledge, we are the first to apply gradient-based distribution shift detection specifically to trajectory prediction. Unlike previous methods that primarily focus on classification tasks, our approach addresses the unique challenges of sequential trajectory data by introducing a self-supervised forecasting task and using gradients from the latent representation space.

\section{Proposed Method}
\label{sec:method}

\begin{figure}[t]
  \centering
  \includegraphics[width=\linewidth]{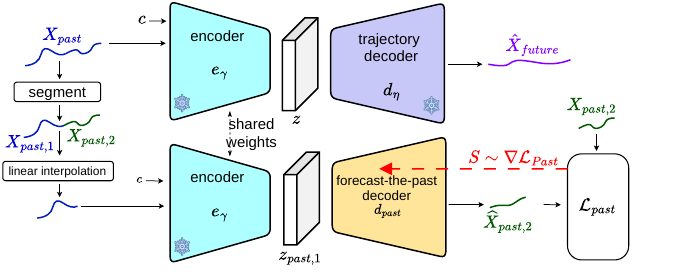}
  \caption{Overview of our post-hoc gradient-based distribution shift detection method.
  The upper part shows the pre-trained encoder-decoder for trajectory prediction.
  Given its encoder $e_{\gamma}$, we propose our forecast-the-past decoder $d_{past}$ for distribution shift detection.
  The historical trajectory $X_{past}$ is split into two segments, $X_{past,1}$ and $X_{past,2}$, which are resized via linear interpolation.
  The frozen encoder processes $X_{past,1}$ to produce the latent representation $z_{past,1}$, while our forecast-the-past decoder $d_{past}$ predicts $X_{past,2}$.
  The gradient of the forecasting loss $\mathcal{L}_{past}$ with respect to the input of the decoder's last layer $h_{last}$ serves as the distribution shift score $S = \nabla \mathcal{L}_{past}$.}
  \label{fig:method}
\end{figure}

Let $X_{past} = \{x_1, x_2, \dots, x_n\}$ be the observed past trajectory of an agent with $x_i \in\mathbb{R}^2$ corresponding to its 2D bird's-eye view coordinates over $n$ time steps, and let $c$ be the surrounding scene context. The primary task is to predict the future trajectory $X_{future} = \{x_{n+1}, \dots, x_N\}$, which is defined as the set of coordinates over $N$ future time steps. To this end, the method assumes access to a pre-trained trajectory prediction model, composed of an encoder $e_\gamma$ and decoder $d_\eta$. The encoder maps the past trajectory $X_{past}$ and scene context $c$ to the latent representation $z = e_\gamma(X_{past}, c)$. From this latent vector, the decoder explicitly outputs the parameters of a Gaussian mixture model to predict the multi-modal distribution of future trajectories $\hat{X}_{future} = d_\eta(z)$. This distribution is defined as $\hat{X}_{future} \sim \sum_{k=1}^{K} \pi_k \mathcal{N}_k(\mu_k, \Sigma_k)$ where $k$ is the number of mixture components, $\pi_k$ defines the component probabilities, and $\mu_k$ and $\Sigma_k$ are the respective means and variances.

\blue{The objective of our method is to predict a distribution shift score $S$ for the observed traffic scene at test time, relying exclusively on the available inputs $X_{past}$ and $c$. A high score $S$ identifies unfamiliar scenes that deviate from the training distribution. Distributional shifts produce distinct gradient magnitudes within the network parameters. However, extracting gradients requires a loss formulation, and the standard trajectory forecasting loss depends on the unavailable future trajectory $X_{future}$. To extract the gradients at test time, the method introduces a self-supervised proxy task. The gradient of this proxy loss yields the target distribution shift score $S$, which directly serves safety-critical tasks such as anomaly detection and distribution shift filtering.}

\subsection{Forecasting the Past}
\label{subsec:self_supervised_anomaly}

Models trained on normal data exhibit different gradient magnitudes when processing inputs from a different distribution \cite{huang2021importance,chen2018gradnorm}. Since the standard trajectory forecasting loss requires future trajectories $X_{future}$, which are unavailable, we introduce a surrogate task: \textit{Forecast the Past}, i.e.~predict the second half of the past trajectory, given the first half of the trajectory and the scene context. Figure~\ref{fig:method} provides an overview of our approach.

\begin{figure*}
  \includegraphics[width=\linewidth]{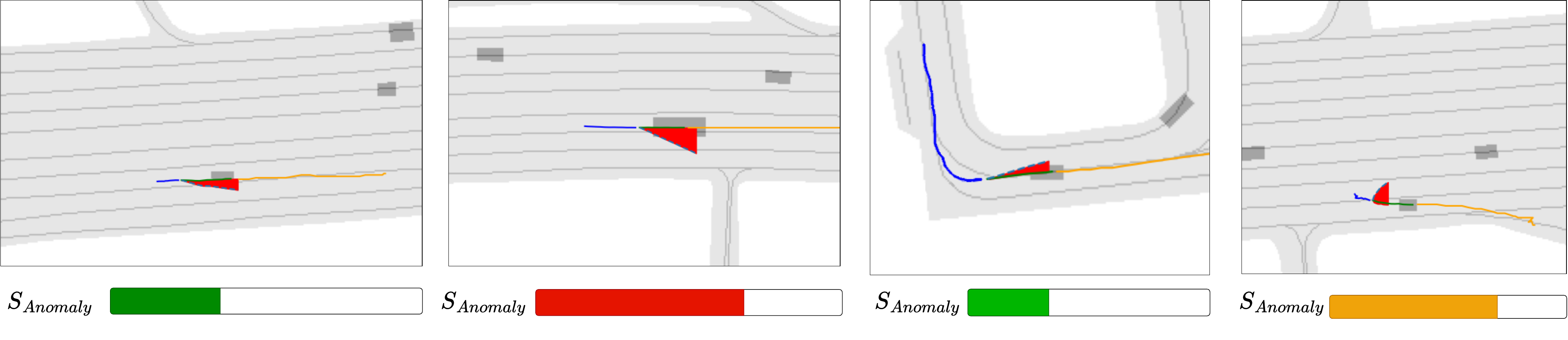}
  \caption{Qualitative example of our gradient-based distribution shift detection method. The figure shows trajectory samples from in-distribution and distribution shifted scenarios. For each sample, we display the historical trajectory split into two segments ($X_{past,1}$ with the blue line and $X_{past,2}$ with the green line), the forecasted second segment using our self-supervised decoder $\hat{X}_{past,2}$ with a dashed line, and the corresponding gradient-based anomaly score $S$. Higher gradient magnitudes indicate distribution shift samples that deviate from learned motion dynamics.}
  \label{fig:qualitative_sample}
\end{figure*}

We utilize the feature representations from the pre-trained encoder $e_\gamma$ for a self-supervised forecasting task on the historical data. First, we partition the historical trajectory $X_{past}$ into two contiguous, equal-length segments: an early history $X_{past,1} = \{x_1, \dots, x_{n/2}\}$ and a later history $X_{past,2} = \{x_{n/2+1}, \dots, x_n\}$ (Figure~\ref{fig:qualitative_sample}).
To conform to the dimensional requirements of the pre-trained model, we resample both segments via linear interpolation. Specifically, $X_{past,1}$ is upsampled to the encoder's input length $n$, yielding $\tilde{X}_{past,1}$, while $X_{past,2}$ is resampled to the model's prediction horizon $N-n$, yielding $\tilde{X}_{past,2}$. We then introduce a decoder, $d_{past}$, which is trained to predict the resampled later history from the latent representation of the early history, \ie to map $z_{past,1} = e_\gamma(\tilde{X}_{past,1})$ to $\tilde{X}_{past,2}$.

During training, the parameters of the pre-trained encoder $e_\gamma$ remain frozen. It processes the resampled early history $\tilde{X}_{past,1}$ along with contextual information $c$ to produce a latent representation $z_{past,1} = e_\gamma(\tilde{X}_{past,1}, c)$. Similarly to the primary trajectory forecasting decoder $d_{\eta}$, the decoder $d_{past}$ is modelled as a Mixture Density Network. It predicts a multimodal distribution for the resampled later history, $\tilde{X}_{past,2}$, by outputting the parameters of a Gaussian Mixture Model (GMM) with $K$ components:
\begin{gather*}
p(\tilde{X}_{past,2} | z_{past,1}) = \\
\sum_{k=1}^K \pi_k(z_{past,1}) \mathcal{N}(\tilde{X}_{past,2}; \mu_k(z_{past,1}), \Sigma_k(z_{past,1})),
\end{gather*}
where $\sum_{k=1}^K \pi_k(z_{past,1}) = 1$ and $\pi_k(z_{past,1}) \geq 0$ are the mixture coefficients. We train only the parameters of $d_{past}$ by minimizing the negative log-likelihood (NLL) of the ground-truth trajectory:
\begin{align*}
\mathcal{L}_{\text{past}} &= -\log p(\tilde{X}_{past,2} | z_{past,1}) \\
&= -\log \sum_{k=1}^K  \pi_k(z_{past,1}) \\
&\qquad \mathcal{N}(\tilde{X}_{past,2}; \mu_k(z_{past,1}), \Sigma_k(z_{past,1})).
\end{align*}
Next, we explain how to use $d_{past}$ to detect distributional shifts.

\subsection{Gradient-Based Anomaly Score}
\label{subsec:gradient_anomaly_score}

Once the past forecasting decoder \(d_{past}\) is trained with self-supervision, we can compute meaningful gradients at test time to detect distributional shifts. 
We define our distribution shift score using the gradient with respect to the pre-activation of the final layer of \(d_{past}\). 
Let the decoder \(d_{past}\) be a composition of \(L\) layers: 
\begin{equation}
 d_{past} = f_L \circ f_{L-1} \circ \cdots \circ f_1.
\end{equation}
Let \(h_L\) denote the input to the last layer \(f_L\), such that \(f_L(h_L)\) produces the triplet \((\mu_k(z), \Sigma_k(z), \pi_k(z))\) containing the Gaussian mixture means, covariances, and coefficients. 
The anomaly score \(S\) is defined as the L2 norm of the gradient of the surrogate loss function \(\mathcal{L}_{\text{past}}\) with respect to \(h_{L}\):
\begin{equation}
S(X_{\text{past}},c) = \left\Vert \nabla_{h_{L}}\mathcal{L}_{\text{past}} \right\Vert_{2},
\end{equation}
where the gradient is computed via the chain rule~\cite{rumelhart1986learning}:
\begin{equation}
\nabla_{h_{L}}\mathcal{L}_{\text{past}}
=\frac{\partial\mathcal{L}_{\text{past}}}{\partial(\mu_{k},\Sigma_{k},\pi_{k})} \cdot \frac{\partial(\mu_{k},\Sigma_{k},\pi_{k})}{\partial h_{L}}.
\label{eq:gradient-developed}
\end{equation}

Unlike related methods that rely primarily on feature space representations or loss magnitudes \cite{yao2024trajoutofdistribution, wiederer2023joint, pmlr-v164-wiederer22a, ahmadi2024curb}, this approach extracts the anomaly score from the gradient space. 
As demonstrated by ElAraby et al.~\cite{elarabygrood}, the gradients capture richer discriminative information than raw feature distances or output confidence scores. 
By capturing the interaction between the loss landscape and the internal representation of the model (Eq.~\ref{eq:gradient-developed}), the gradient encodes how strongly the network hidden representation must be updated to fit a given input.   
Consequently, anomalous samples induce distinctly larger and more erratic gradient responses compared to stable, in-distribution data.
We show empirically in section~\ref{subsec:loss_function_ood} a direct comparison between gradient-based scores and output-based scores for distribution shift detection.
Next, we discuss our experiments in more detail.

\section{Experiments}
\label{sec:result}

We evaluate our gradient-based distribution shift detection method on two datasets and in the simulator. 
The datasets benchmark different types of distribution shifts, \ie, the first contains environmental distribution shifts (weather, city, time of the day) while the second contains shifts in motion behaviour. 
In the simulator, we show that our method works in an online environment for monitoring failures of a planning policy.

\subsection{Experimental Setup}
\label{subsec:experimental_setup}

The following introduces the datasets and the evaluation protocol.

\paragraph{Datasets}
\label{par:datasets}

We utilize the Shifts Vehicle Motion Prediction Dataset~\cite{malinin2021shifts}, which is designed to evaluate trajectory prediction under distribution shifts for automated driving. 
This dataset contains \SI{388406} and \SI{36 804} sequences for training and testing, respectively, collected across six locations (Moscow, Skolkovo, Innopolis, Ann Arbor, Modiin, and Tel Aviv), three seasons (Summer, Autumn, Winter), three times of day (Astronomical Night, Daylight, Twilight), and four weather conditions (No precipitation, Rain, Sleet, Snow). 
The distribution shifts are \textit{environmental}, where the testing set contains cities that are unseen during training (\ie, Ann Arbor and Tel Aviv) and additional precipitation conditions (\ie, Rain, Sleet, Snow).
Each scene spans 10 seconds, divided into 5 seconds of history and 5 seconds of ground truth future for prediction.
Additionally, we adapt the Argoverse 1 motion forecasting dataset~\cite{Argoverse} for distribution shift detection by artificially creating \textit{behavioural} out-of-distribution scenarios.
Following the taxonomy of Schmidt et al.~\cite{schmidt2022meat}, we remove specific trajectory manoeuvrers from the training set: left turns, right turns, and trajectories exceeding maximum velocity thresholds.
We use the removed manoeuvres as the out-of-distribution scenarios during testing.
The behaviour clustering methods are described in Sections~\ref{par:traj_direction}.

\paragraph{Evaluation Protocol}
\label{par:evaluation_protocol}

\begin{figure*}
  \centering
  \begin{subfigure}{0.32\linewidth}
    \centering
      \includegraphics[width=0.7\linewidth]{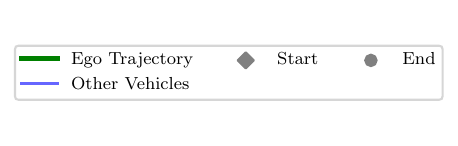}
      \includegraphics[width=0.9\linewidth]{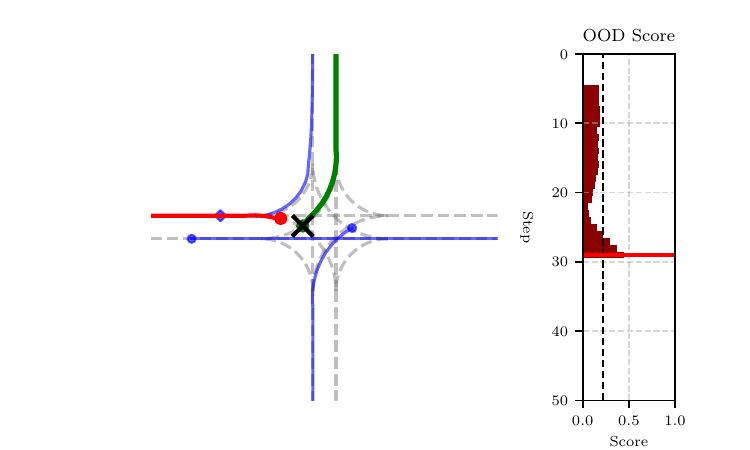}
      \caption{Merge Crash}
      \label{fig:hw_merge_crash}
  \end{subfigure}
  \hfill {\color{lightgray}\vrule} \hfill
  \begin{subfigure}{0.32\linewidth}
    \centering
    \includegraphics[width=0.7\linewidth]{img/highway_samples/ego_traj_legend.pdf}
    \includegraphics[width=0.9\linewidth]{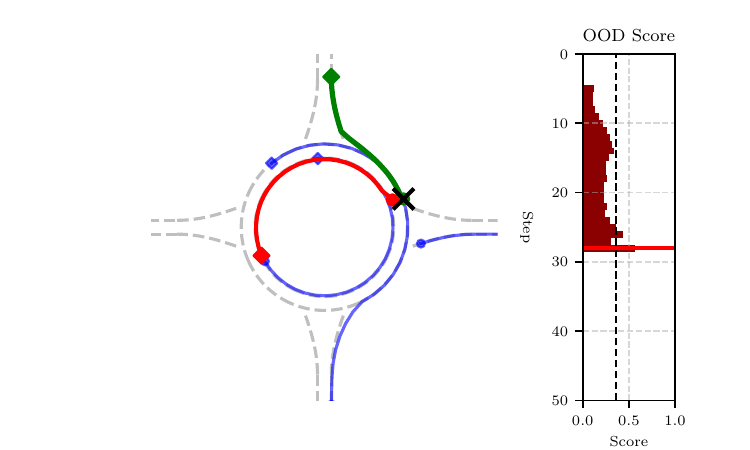}
    \caption{Roundabout Crash}
    \label{fig:hw_roundabout_crash}
  \end{subfigure}
  \hfill {\color{lightgray}\vrule} \hfill
  \begin{subfigure}{0.32\linewidth}
    \centering
    \includegraphics[width=0.7\linewidth]{img/highway_samples/ego_traj_legend.pdf}
    \includegraphics[width=0.9\linewidth]{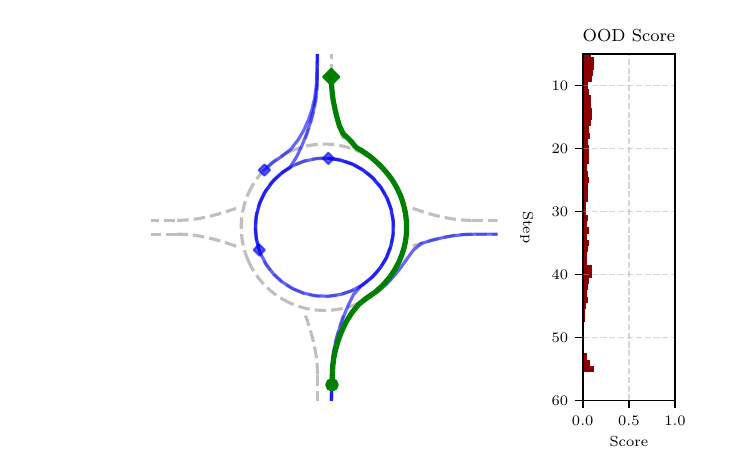}
    \caption{Roundabout Normal}
    \label{fig:hw_roundabout_normal}
  \end{subfigure}

  \caption{Highway simulator scenarios depicting a merge crash, a roundabout crash, and normal roundabout navigation. We mark the start and end points of trajectories, highlight the colliding vehicle in a different color, and mark the crash point. A horizontal red line in the OOD score plots indicates the crash timestep, and a thin line denotes the anomaly threshold. Collisions in Figures~\ref{fig:hw_merge_crash} and \ref{fig:hw_roundabout_crash} produce high gradient values, while normal behavior in Figure~\ref{fig:hw_roundabout_normal} shows gradients below the threshold. In Figure~\ref{fig:hw_merge_crash}, the method predicts stopping (black line), and the score increases immediately after step 20, successfully detecting the collision. In Figure~\ref{fig:hw_roundabout_crash}, the gradient score reaches a peak at timestep 26, preceding a collision after 4 steps. Figure~\ref{fig:hw_roundabout_normal} shows normal driving behavior in a roundabout, with a stable gradient score over time.}
  \label{fig:highway_samples}
\end{figure*}

For evaluation on Shifts~\cite{malinin2021shifts} (Figure~\ref{fig:highway_samples}), we compare our method with the baseline provided by the Shifts dataset, \ie the RNN-based behavioral cloning network (RIP-BC)~\cite{codevilla2018end}, the autoregressive flow–based deep imitative model (RIP-DIM)~\cite{Rhinehart2020Deep} and the latent Gaussian mixture model (lGMM)~\cite{wiederer2023joint}. 
To train our model, we take the pre-trained encoder-decoder network from \cite{wiederer2023joint}, and train our self-supervised decoder on top of the encoder.
For the experiments on Argoverse~\cite{Argoverse}, we use the HiVT trajectory prediction model~\cite{zhou2022hivt} as our encoder-decoder.
We train individual HiVT predictors for each training set (turn left, turn right and max velocity) and evaluate their performance on the full Argoverse validation set.
We compare our approach with several baselines, including one-class support vector machine (OC-SVM), isolation forest (IF) and kernel density estimation (KDE), all of which trained on the latent space $z = e_\gamma(X_{past})$ of HiVT~\cite{zhou2022hivt}.

\paragraph{Evaluation Metrics}
\label{par:evaluation_metrics}

Like the prior work~\cite{malinin2021shifts, wiederer2023joint}, we report the area under the receiver operating characteristic curve (AUROC) as the metric for distribution shift detection.

\subsection{Implementation Details}
\label{subsec:implementation_details}

\paragraph{Trajectory Segmentation}
For the self supervised task we split the historical trajectory as follows: the first half $X_{past,1}$ (12 timesteps for Shifts~\cite{malinin2021shifts} and 10 timesteps for Argoverse \cite{Argoverse}) is used to train the self-supervised decoder $d_{past}$ and the second half $X_{past,2}$ (13 timesteps for Shifts \cite{malinin2021shifts} and 10 timesteps for Argoverse \cite{Argoverse}) is used as ground truth trajectory. Before feeding the trajectories to the model, we expand $X_{past,1}$ and $X_{past,2}$ to the size of original trajectory $X_{past}$ (25 for Shifts and 20 for Argoverse) and $X_{forecast}$ (25 for Shifts and 30 Argoverse) using linear interpolation to keep dimensionality consistency with the original encoder $e_\gamma$.

\paragraph{Trajectory Manoeuvrer Clustering}
\label{par:traj_direction}
For the Argoverse evaluation (Table~\ref{tab:hivt_results}) we cluster the trajectory orientation using a computational geometry technique. 
Given consecutive points $P_i, P_{i+1}, P_{i+2}$ in trajectory $T$, we compute the 2D cross product:
\begin{equation}
\phi_i = \Delta p_{i,x} \Delta p_{i+1,y} - \Delta p_{i,y} \Delta p_{i+1,x}
\end{equation}
where $\Delta p_i = P_{i+1} - P_i$ and $\Delta p_{i+1} = P_{i+2} - P_{i+1}$. Positive values indicate left turns, negative values indicate right turns, and zero indicates straight motion. 
The overall trajectory orientation is $\phi_{total} = \sum_{i=1}^{N-2} \phi_i$. 
For distribution shift detection, we remove trajectories with $\phi_{total}<1$ from the training set in the experiment "turn\_right" and $\phi_{total}>-1$ in the experiment "turn\_right". We choose these threshold values to ensure that the model learns a strong directional bias.

\begin{figure}
  \centering
  \begin{subfigure}{0.48\linewidth}
      \includegraphics[width=\linewidth]{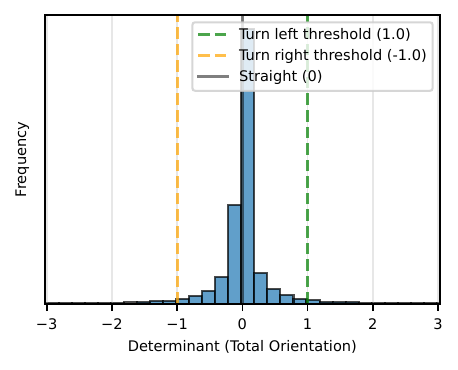}
      \caption{Distribution of rotation determinant in Argoverse V1.}
      \label{fig:hist_argoverse_determinant}
  \end{subfigure}
  \hfill
  \begin{subfigure}{0.48\linewidth}
      \includegraphics[width=\linewidth]{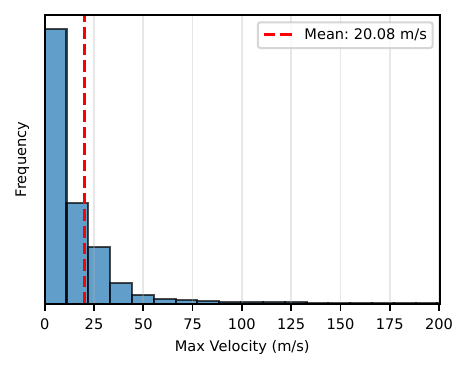}
      \caption{Distribution of max velocity in Argoverse V1.}
      \label{fig:hist_argoverse_velocity}
  \end{subfigure}
  \caption{Distribution of rotation determinant and max velocity in Argoverse V1. Further details about their calculation are in Paragraphs~\ref{par:traj_direction} and \ref{par:traj_max_vel}. We observe that trajectories are skewed towards turning right in Argoverse \cite{Argoverse}.}
  \label{fig:hist_argoverse_shifts}
\end{figure}

\paragraph{Trajectory Velocity Clustering}
\label{par:traj_max_vel}

For the maximum velocity experiment in Argoverse~\cite{Argoverse}, we compute velocity from the trajectory coordinates. 
The Argoverse dataset samples trajectory points at 10 Hz, providing consecutive positions $P_i = (x_i, y_i)$ and $P_{i+1} = (x_{i+1}, y_{i+1})$ with time interval $\Delta t = 0.1$ seconds. 
We calculate instantaneous velocity as:
\begin{equation}
v_i = \frac{\sqrt{(x_{i+1} - x_i)^2 + (y_{i+1} - y_i)^2}}{\Delta t}
\end{equation}
The maximum velocity for a trajectory is $v_{max} = \max_i v_i$.
For distribution shift detection, we remove trajectories with $v_{max} > v_{threshold}$ from the training set, where $v_{threshold}$ is set to the median of observed velocities in the training data.

\paragraph{Training and Architecture}
\label{par:training-architecture}
For Argoverse \cite{Argoverse} and Shifts \cite{safeshift}, we freeze the encoder trained on the forecasting and we train a decoder with the same architecture and hyper-parameters of the trajectory predictor. For the Highway simulator we train a Transformer \cite{vaswani2017attention} encoder and MLP decoder for \textit{intersection} and \textit{merge}. We train all the models with Adam optimizer and learning rate $1e-4$.

\begin{table}[t]
  \centering
  \caption{Performance of Distribution shifts detection on Shifts \cite{malinin2021shifts}}
  \label{tab:shifts_results}
 
  \begin{tabular}{lc}
  \toprule
  \textbf{Model} & \textbf{AUROC (\%)} \\
  \midrule
  RIP-BC (K=1)~\cite{safeshift,codevilla2018end} & 52.8\%  \\
  RIP-BC (K=5)~\cite{safeshift,codevilla2018end} & 52.1\%  \\
  RIP-DIM (K=1)~\cite{safeshift,Rhinehart2020Deep} & 51.8\%  \\
  RIP-DIM (K=5)~\cite{safeshift,Rhinehart2020Deep} & 50.9\%  \\
  lGMM \cite{wiederer2023joint} & 56.8\%  \\
  Our Method & \textbf{71.0}\%  \\
  \bottomrule
  \end{tabular}

\end{table}

\begin{table}[t]
  \caption{OOD detection performance in terms of AUROC on the Argoverse Dataset.}
  \resizebox{\linewidth}{!}{%
  \begin{tabular}{lcccccc}
    \toprule
    & \textbf{KDE} & \textbf{OC-SVM} & \textbf{IF}  & \textbf{Our Method} \\
    \midrule
    Turn right & 56.1\% & 56.4\% & 61.0\%  & \textbf{71.3}\%\\
    Turn left & 55.2\% & 52.8\% & 53.2\% & \textbf{70.3}\% \\
    Max Velocity & 70.8\% & 61.6\% & 65.4\% & \textbf{79.2}\% \\
    \bottomrule
  \end{tabular}
  }
  \label{tab:hivt_results}
\end{table}

\begin{figure}
  \centering
  \begin{subfigure}{0.48\linewidth}
      \includegraphics[width=\linewidth]{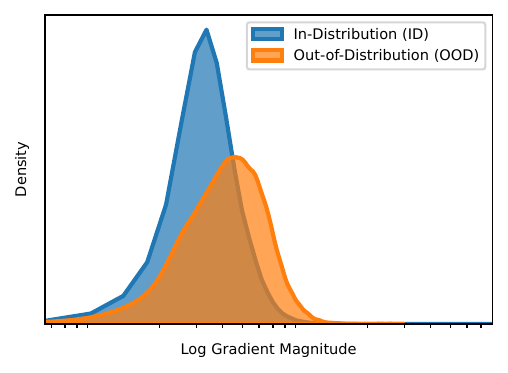}
      \caption{Shifts \cite{malinin2021shifts} dataset gradient distributions.}
      \label{fig:shifts_gradients}
  \end{subfigure}
  \hfill
  \begin{subfigure}{0.48\linewidth}
      \includegraphics[width=\linewidth]{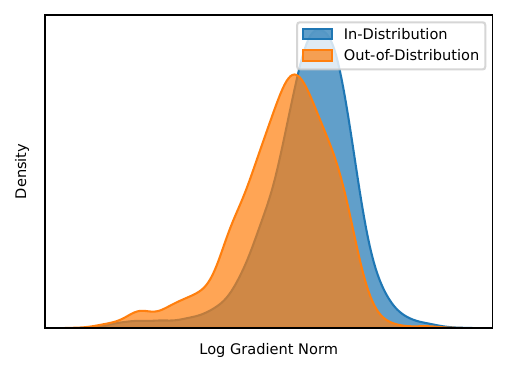}
      \caption{Argoverse dataset gradient distributions.}
      \label{fig:argoverse_gradients}
  \end{subfigure}
  \caption{Gradient Distribution for In-Distribution vs. Out-Of-Distribution samples for Shifts~\cite{malinin2021shifts} and Argoverse \cite{Argoverse}. We observe different distribution for ID (Blue area) and OOD (Red Area) in Shifts Dataset~\cite{malinin2021shifts}.}
  \label{fig:joodu_gradients}
\end{figure}

\begin{figure}
  \centering
  
  \footnotesize{Highway simulator gradient distributions.}
  \vspace{0.01cm}
  \includegraphics[width=0.5\linewidth]{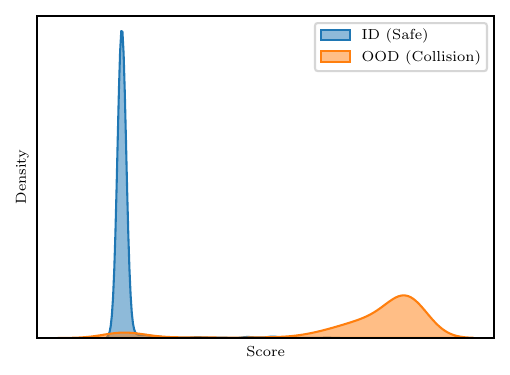}
  \caption{\blue{Kernel density estimation (KDE) \cite{chen2017tutorial} of the last-layer gradients in the Highway environment~\cite{highway-env} for the \textit{intersection} driving task. We observed very distinct gradients between ID on OOD samples, leading to almost perfect collision detection.}}
  \label{fig:activation_mlp_decoder}
\end{figure}

\begin{table}[ht]
  \centering
  \caption{\blue{Ablation study of model architectures and scoring strategies for distribution shift detection on the Shifts dataset~\cite{malinin2021shifts} using the \textit{AUROC} metric. We compare standard and masked autoencoders against our method using different scores: gradients w.r.t. the last layer (\textbf{last}), the latent space (\textbf{latent}), all model parameters (\textbf{all}), as well as the self-supervised loss function of the models (\textbf{loss}).}}
  \label{tab:autoencoder_masked_ablation}
\begin{tabular}{lcccc}
  \toprule
  Model & \textbf{last} & \textbf{lat.} & \textbf{all} & \textbf{loss} \\
  \midrule
  AE         & 50.56\% & 56.26\% & 57.29\% & 50.00\% \\
  MAE        & 50.51\% & 56.08\% & 57.46\% & 48.95\% \\
  Our method & \textbf{71.30\%} & \textbf{60.20\%} & \textbf{58.70\%} & \textbf{51.00\%} \\
  \bottomrule
\end{tabular}
\end{table}

\subsection{Results Analysis}
\label{subsec:results_analysis}

Our experimental results show different patterns in gradient-based distribution shift detection across different datasets. 

\paragraph{Shifts Results}
Table~\ref{tab:shifts_results} demonstrates our method's substantial improvement on the Shifts dataset~\cite{malinin2021shifts}, achieving $\sim$71.0\% ROC AUC compared to the baseline's 56.8\%. This significant performance gain indicates that our gradient-based approach effectively captures environmental distribution shifts such as new cities and weather conditions. As shown in Fig.~\ref{fig:roc_curve_shifts}, reconstruction-based methods cannot capture well anomalies, as the reconstruction loss focuses only on the reconstructed trajectory.

\paragraph{Argoverse Results}
Table~\ref{tab:hivt_results} shows strong performance across all behavioral anomaly detection tasks. The method achieves 71.3\% and 70.3\% AUROC for right and left turn detection, respectively, outperforming KDE (55.2-56.1\%), OC-SVM (52.8-56.4\%), and Isolation Forest (53.2-61.0\%). The performance gap between turn directions is attributed to the skewness of the determinant distribution (Fig.~\ref{fig:hist_argoverse_determinant}). For velocity-based anomalies, the method achieves 79.2\% AUROC, exceeding the best baseline of 70.8\% from KDE.

\paragraph{Gradient Pattern Analysis}
Figure~\ref{fig:joodu_gradients} shows distinct gradient behavior across datasets. On Shifts~\cite{malinin2021shifts}, ID and OOD distributions are clearly separated, with OOD samples producing larger gradient norms. This matches the expected behavior for anomalous inputs in steeper regions of the loss landscape.

On Argoverse, OOD samples often show lower gradient norms than ID samples. We hypothesize from our experiments that this failure is due to training failure convergence, since removing a large portion of trajectories lead to high loss function at the end of the training. 

\paragraph{Stability of Results}
Across repeated runs, the results on Shifts~\cite{malinin2021shifts} and Highway remain statistically stable, with consistent ranking and separation between ID and OOD samples. In contrast, Argoverse shows noticeably higher variance and less reliable estimates. We attribute this behavior to unstable training dynamics under the maneuver-filtered setup. The same effect is reflected in the inverted gradient-scale relation on Argoverse, where OOD samples can yield lower gradient norms than ID samples (Fig.~\ref{fig:joodu_gradients}), which indicates reduced robustness of the score in this setting.

\begin{table}[ht]
  \centering
  \caption{\blue{Distribution Shifts detection performance (AUROC) across different scenarios in the Highway~\cite{highway-env} Environment.}}
  \label{tab:highway_results}
  \resizebox{\linewidth}{!}{%
\begin{tabular}{lccc}
    \toprule
    \textbf{Method} & \textbf{Merge} & \textbf{Intersection} & \textbf{Roundabout} \\
    \midrule
    AE & 99.9\% & 64.3\% & 85.1\% \\
    Masked AE & 99.9\% & 65.5\% & 85.8\% \\
    IF & 99.1\% & 67.9\% & 91.1\% \\
    Our method & \textbf{99.9\%} & \textbf{96.1\%} & \textbf{95.8\%} \\
    \bottomrule
\end{tabular}%
}
\end{table}

\begin{figure}
  \centering
    \footnotesize{Last layer Highway~\cite{highway-env} Roundabout Inputs}
    \vspace{0.1cm}
  
    \includegraphics[width=0.7\linewidth]{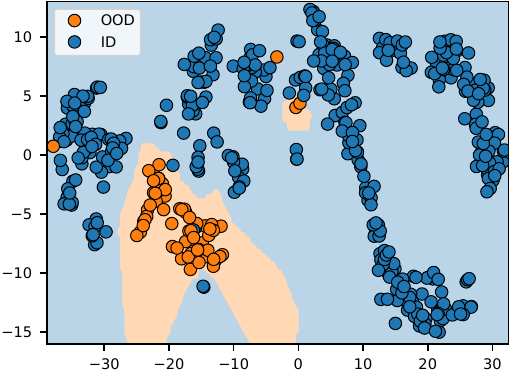}
    \caption{\blue{t-SNE visualization of the decoder's last layer inputs for the Highway~\cite{highway-env} Roundabout scenario. Safe (ID) and collision (OOD) trajectories form well-separated clusters, demonstrating the discriminative power of the learned representations.}}
    \label{fig:kde_mlp}
\end{figure}

\subsection{Early-Detection of Planning Failures}
  
  To showcase the practicality of the approach, we evaluate it to identify planning failures in online simulation. We define multiple driving tasks (roundabout, merge, and intersection) in the Highway environment~\cite{highway-env} and train a reinforcement learning policy using proximal policy optimization (PPO) \cite{schulman2017proximal} from Stable Baselines 3 \cite{stable-baselines3} for each task. Operating at a control frequency of 5 Hz, the trained PPO policy generates a training dataset comprising 10,000 safe, collision-free episodes. Given this data, we train a predictive model to \textit{forecast the past} i.e. predict the last 5 time-steps of the trajectory using the Transformer encoder and an MLP decoder architecture, inspired by HiVT \cite{zhou2022hivt} and described in Section~\ref{subsec:implementation_details}.
  
  For the evaluation, we deploy the method in an online fashion to monitor the planning policy across a balanced test set of 1,000 safe and 1,000 collision episodes. We compute the distribution shift scores using the last-layer gradient as detailed in Section~\ref{subsec:gradient_anomaly_score}. Crucially, in the collision episodes, we capture the score one second before the actual impact occurs, testing the method's ability to provide an early warning. We compare the performance against standard self-supervised autoencoder, masked autoencoder, and IF baselines (Section~\ref{subsec:experimental_setup}) trained on the identical dataset. We omit KDE and OC-SVM baselines due to severe scalability bottlenecks related to kernel estimation during online operation.

  \paragraph{Detection Results}
\label{par:highway-results}

Table~\ref{tab:highway_results}, along with Figure~\ref{fig:kde_mlp} and Figure~\ref{fig:activation_mlp_decoder}, present the method performance in the interactive Highway~\cite{highway-env} environment across three planning tasks: \textit{merge}, \textit{intersection}, and \textit{roundabout}.
Baseline anomaly scoring functions are configured correctly so that higher scores target the collision class, resolving prior inverted AUROC results.
The gradient-based method and reconstruction baselines (Autoencoder and Masked Autoencoder) achieve a near-perfect 99.9\% AUROC on the \textit{merge} scenario.
This high accuracy occurs because the \textit{merge} scenario presents distinct, easily separable driving patterns between normal and collision events.
However, in more complex environments, the proposed gradient-based score substantially outperforms the baselines.
On the \textit{intersection} task, the gradient-based method achieves 96.1\% AUROC, compared to 64.3\% for the Autoencoder and 65.5\% for the Masked Autoencoder.
Similarly, the \textit{roundabout} scenario yields 95.8\% AUROC for the proposed method, exceeding the Autoencoder (85.1\%) and Masked Autoencoder (85.8\%).
While Isolation Forest also yields competitive results on the \textit{merge} (99.1\%) and \textit{roundabout} (91.1\%) tasks, the gradient-based score consistently surpasses all baselines across evaluated configurations.

\paragraph{Runtime evaluation}
\label{par:performance-early-planning}
Since we are testing in an online simulator, we evaluate the computational cost of our approach. While the PPO policy inference requires $\sim 1$ ms per step, our method computes the anomaly score in $\sim 3$ ms using the Transformer encoder with MLP decoder, and $\sim 4$ ms with the full Transformer encoder-decoder. Although our implementation is unoptimized compared to the established Stable Baselines 3~\cite{stable-baselines3} PPO implementation~\cite{schulman2017proximal}, the runtime remains suitable for real-time monitoring. Theoretically, the computational overhead is minimal, as the method primarily involves a forward pass and a single gradient computation with respect to the last layer. Given a 10 Hz operating frequency (100 ms per cycle), this minimal overhead reserves the remaining 96 ms entirely for the upstream perception and localization modules.

\subsection{Ablation Studies}
\label{subsec:ablation_studies}

\begin{figure}
  \centering
  \includegraphics[width=0.7\linewidth]{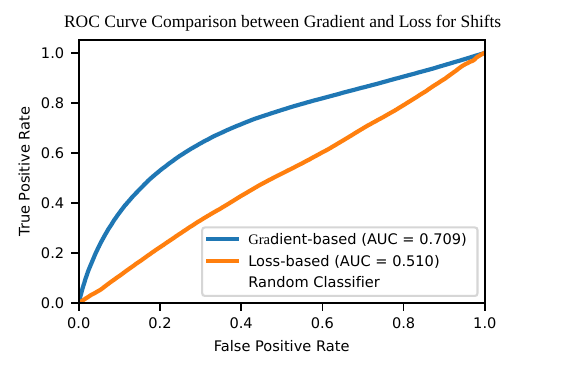}
    \caption{Gradient vs loss-based OOD detection on the Shifts dataset~\cite{malinin2021shifts}. While the loss function $\mathcal{L}_{past}$ reaches 51.0 \% AUROC, the gradient of the loss function $\nabla \mathcal{L}_{past}$ reaches 70.9 \% AUROC showcasing the effectiveness of gradient-based OOD detection (Sec.~\ref{subsec:gradient_anomaly_score}).}
    \label{fig:roc_curve_shifts}

\end{figure}

\paragraph{Loss Function for Distribution Shift Detection}
\label{subsec:loss_function_ood}

To prove the validity of our method we tested our Distribution Shift detection method on Shifts~\cite{malinin2021shifts} using the loss function value $L_{past}$ as score instead of the gradient of the function $ || \nabla_{h_L} \mathcal{L}_{past} ||_2 $ w.r.t. the last hidden layer input $h_L$, as described in Sec.~\ref{subsec:gradient_anomaly_score}.
\begin{blueenv}
As demonstrated in Fig.~\ref{fig:roc_curve_shifts} and Table~\ref{tab:autoencoder_masked_ablation}, using the loss function does not provide significant distribution shift information, yielding an AUROC of approximately 50\%. In contrast, the gradient-based approach effectively captures variations in the loss landscape and performs significantly better for distribution shift detection. This confirms our hypothesis from Sec.~\ref{subsec:gradient_anomaly_score} showing the superiority of the gradients in terms of Distribution Shifts detection versus the loss space.   
\end{blueenv}

\begin{blueenv}
  \paragraph{Comparison with Other Self-Supervised Approaches}
  \label{par:selfsupervised}
  Table~\ref{tab:autoencoder_masked_ablation} compares standard self-supervised methods for distribution shift detection on the Shifts dataset~\cite{malinin2021shifts}: an \textit{Autoencoder} (AE) that reconstructs the full trajectory, and a \textit{Masked Autoencoder} (MAE) that predicts randomly masked trajectory states. 
  Formally, the AE minimizes the mean squared error (MSE) of the full historical trajectory: $\mathcal{L}_{AE} = (X_{past}, AE(X_{past}))^2$. 
  The MAE processes a masked input, $\bar{X}_{past} = M \odot X_{past}$, where $M$ is a random binary mask, and minimizes $\mathcal{L}_{MAE} = ( (1- M) \odot X_{past}, MAE(\bar{X}_{past}))^2$.
  Our method, \textit{forecasting the past}, applies causal masking by estimating the trajectory's second half. 
  For a fair comparison, the masked autoencoder also masks exactly 50\% of the trajectory steps.
  We compute anomaly scores using the loss function magnitude (\textbf{recon}/\textbf{pred error}), and the L2 norm of the loss gradient with respect to all parameters (\textbf{all}), the latent space (\textbf{latent}), and the final layer (\textbf{last}).
  The results indicate that the loss magnitude poorly identifies distribution shifts, yielding approximately 50\% AUROC across all models. 
  Gradient-based scores consistently perform better. 
  Notably, our causal forecasting method combined with the \textbf{last} layer gradient achieves 71.30\% AUROC, outperforming reconstruction and random masking methods by over 20\%.
\end{blueenv}

\section{Conclusion}
\label{sec:conclusion}

We presented a post-hoc gradient-based method for distribution shift detection in trajectory prediction that does not affect the original model forecasting performance. A self-supervised decoder is trained to forecast the second half of the historical trajectory from the first half, providing a proxy task for computing gradients at test time. The L2 norm of this forecasting loss gradient with respect to the decoder final layer serves as the anomaly score, outperforming both loss-based and reconstruction-based baselines. On the Shifts dataset, the method achieves 71.0\% AUROC compared to 56.8\% for existing approaches, with consistent improvements on Argoverse and in the interactive Highway simulator for early detection of planning failures. The results establish gradient-based detection as a promising direction for robust trajectory prediction in autonomous driving.

\section*{Acknowledgement}
The research leading to these results is funded by the German Federal Ministry for Economic Affairs and Energy within the project “NXT GEN AI METHODS – Generative Methoden für Perzeption, Prädiktion und Planung". The authors would like to thank the consortium for the successful cooperation.

{
    \small
    \bibliographystyle{ieeenat_fullname}
    \bibliography{main}
}

\end{document}